\documentclass{article} 
\usepackage{iclr2024_conference,times}

\usepackage{amsmath,amsfonts,bm}

\def\eqref#1{equation~\ref{#1}}

\def\1{\bm{1}}

\DeclareMathAlphabet{\mathsfit}{\encodingdefault}{\sfdefault}{m}{sl}
\SetMathAlphabet{\mathsfit}{bold}{\encodingdefault}{\sfdefault}{bx}{n}

\usepackage{hyperref}
\usepackage{url}
\usepackage{bbm}
\usepackage{graphicx}
\usepackage{caption}
\usepackage{subcaption}
\usepackage{wrapfig}
\usepackage{booktabs}
\usepackage{authblk}

\title{Calibrating Bayesian UNet++ for Sub-seasonal Forecasting}

\author[1\thanks{Corresponding author: asan18@itu.edu.tr}]{\textbf{Büşra Asan}}
\author[2]{\textbf{Abdullah Akgül}}
\author[3]{\textbf{Alper Ünal}}
\author[2]{\textbf{Melih Kandemir}}
\author[1]{\textbf{Gözde Ünal}}
\affil[1]{Department of Computer Engineering, Istanbul Technical University}
\affil[2]{Department of Mathematics and Computer Science, University of Southern Denmark}
\affil[3]{Eurasia Institute of Earth Sciences, Istanbul Technical University}

\iclrfinalcopy 
\begin{document}

\maketitle

\begin{abstract}
Seasonal forecasting is a crucial task when it comes to detecting the extreme heat and colds that occur due to climate change. Confidence in the predictions should be reliable since a small increase in the temperatures in a year has a big impact on the world. Calibration of the neural networks provides a way to ensure our confidence in the predictions. However, calibrating regression models is an under-researched topic, especially in forecasters. We calibrate a UNet++ based architecture, which was shown to outperform physics-based models in temperature anomalies. We show that with a slight trade-off between prediction error and calibration error, it is possible to get more reliable and sharper forecasts. We believe that calibration should be an important part of safety-critical machine learning applications such as weather forecasters.

\end{abstract}

\section{Introduction}



Seasonal forecasting is a crucial task when it comes to foreseeing the effects of climate change, especially in making predictions and decisions based on these effects. Generating accurate seasonal and sub-seasonal forecasts demands substantial resources, such as the curation of Coupled Model Intercomparison Projects (CMIP) datasets. These datasets combine outputs from over a hundred climate models worldwide, facilitating top-tier climate simulations. Leveraging the vast data reservoirs from CMIP6 \citep{eyring2016}, the latest phase of CMIP, there are ongoing efforts to harness deep learning methodologies for enhanced climate forecasting. For instance, \cite{luo2022bayesian} use Bayesian Neural Networks (BNN) with CMIP6 for climate prediction in the North Atlantic, and \cite{anochi2021machine} use CMIP6 to assess precipitation. \cite{andersson2021seasonal} forecast the change in Arctic sea ice area with the same dataset. In this work, we also utilize the CMIP6 dataset to produce well-calibrated and sharp forecasts which are crucial for climate sciences \citep{gneiting2007probabilistic}. 

We expand the capabilities of the forecast model introduced in \cite{unal_asan_sezen_yesilkaynak_aydin_ilicak_unal_2023} using the calibration approach proposed by \cite{kuleshov2018accurate}. This model is shown to achieve better performance than physics-based methods on sub-seasonal forecasting, especially at predicting temperature anomalies that indicate extreme hot and cold temperatures that are crucial for climate change.

In this work, we calibrate the Bayesian version of the forecaster in a regression setting. We show that the BNNs produce the most calibrated and sharp forecasts. We compare the performance of BNNs, Monte Carlo Dropout (MC-Dropout) \cite{gal2016dropout}, and Deep Ensemble \cite{lakshminarayanan2017simple} methods for assessing climate forecast uncertainty and sharpness, exploring the potential for improved reliability through calibration. Our contributions can be listed as follows:

\begin{enumerate}
    \item We apply calibration to a sub-seasonal forecaster that is able to predict extreme events better than simulations. We show that calibrating deep learning models should be a crucial step while applying deep learning to climate sciences.
    \item We show that well-calibrated forecasters not only produce better confidence intervals but may also improve the sharpness of the forecasts.
    \item This method may be generalized to any other application in climate sciences that gives critical importance to the reliability of the results such as extreme events, precipitation, and natural disasters such as earthquakes, floods, and drought.
\end{enumerate}

\section{Methodology}\label{methods}

We formulate the problem as predicting the monthly average air temperature at 2 meters above the earth surface ($2m$) for each coordinate in a 2D temperature grid which we will name as the temperature map. Our aim is to construct a reliable confidence interval for each coordinate since Bayesian methods often produce uncalibrated results \cite{kuleshov2018accurate}.

We first train the model on CMIP6 climate simulations, then fine-tune it with ERA5 reanalysis data based on real climate measurements. We denote the 2D temperature map at time $t$ as $x_{t}$. Train set $D = \{X_{t}, Y_{t}\}^{T}_{t=1}$ consists of stacked monthly time-series temperature maps as the input. The input $X_{t}$ refers to $x_{t-1:t-k-m}$ which denotes the range of the stacked months and $Y_{t}$ corresponds to $x_{t}$. The periodical month selection process from the given range is described in \cite{unal_asan_sezen_yesilkaynak_aydin_ilicak_unal_2023} and the same setting is used to make a fair comparison. 


\subsection{Bayesian UNet++ for Temperature Prediction}

For temperature forecasting, we convert UNet++ into a BNN \citep{Goan_2020}. BNNs are highly regarded for their capability in quantifying uncertainty, offering robust insights into predictive models \cite{kristiadi2020bayesian}. Thus, we converted the final three layers of the UNet++ architecture into Bayesian convolutional layers, where we model the weights $\theta$ of the neural network as a Gaussian distribution. Letter, we maximize evidence lower bound \cite{kingma2022autoencoding}.



\subsection{Uncertainty Analysis}

\textbf{Confidence Intervals} are used for measuring the uncertainty. Quantiles are calculated from the predictions and checked whether the correct portion of the predictions actually conforms to those intervals. 


\textbf{Calibration} in neural networks refers that if the confidence interval is chosen as $95\%$, then the intervals should capture around $95\%$ of the observed outcomes $Y_{t}$. To measure calibration, we count observations that stay below the predicted upper bound for the quantile $p$ of the sample $t$, then normalize with the size of the dataset. A neural network is said to be calibrated if it satisfies the following
\begin{equation}
    \frac{1}{T}\sum_{t=1}^{T}(\mathbbm{1}\{Y_{t} \leq F_{t}^{-1}(p)\}) \longrightarrow p, 
\label{calib1}
\end{equation}
as $T \rightarrow \infty$ \citep{gneiting2007probabilistic}, where $F_{t}$ refers to the Cumulative Distribution Function (CDF) of the output of the neural network $H$ for the input $X_{t}$. 


\textbf{Calibrated Regression.} We need to match empirical and predicted CDFs according to Equation~\ref{calib1}. Therefore, the training partition of the ERA5 dataset is used to construct a calibration dataset to map the predicted CDF to the empirical CDF. We train a new regressor $R : [0, 1] \rightarrow [0, 1]$ on the calibration dataset. Thus, we expect $R \circ F_{k}$ to be calibrated. CDFs are monotonically increasing functions, hence the choice of $R$ is an Isotonic Regressor \citep{niculescu2005predicting}.

From the training partition $S = \{X_{t-1:t-k-m}, Y_{t}\}^{T'}_{t=1}$ of ERA5, calibration dataset $C= \{ c_{t}, y_{t}\}^{T'}_{t=1}$ is constructed where $c_{t}$ refers to $F_{t}(Y_{t})$ and $y_{t}$ refers to $\hat{P}(F_{t}(Y_{t}))$ using the dataset generation method in \cite{kuleshov2018accurate}. $\hat{P}$ is formulated as
\begin{equation}
    \hat{P} = \frac{1}{T}|\left\{Y_{t} \left| F_{t}\left(Y_{t}\right) < p, t = 1,...,T\right\} \right|
\label{P-hat}
\end{equation}
where $|A|$ refers to the cardinality of the set A. It calculates empirical CDF from the predicted CDF by normalizing the count of output $Y_{t}$ staying below $p^{th}$ quantile of $F_{t}$.
\begin{table}[t!]
\tabcolsep=0pt%
\caption{Metrics (lower the better) for calibrated and uncalibrated versions of the models.}
{
\begin{tabular*}{\textwidth}{@{\extracolsep{\fill}}lcccccc@{}}\toprule%
 & \multicolumn{3}{@{}c@{}}{Uncalibrated}& \multicolumn{3}{@{}c@{}}{Calibrated}
 \\\cmidrule{2-4}\cmidrule{5-7}%
Metrics & CE & MAE & Sharpness
& CE & MAE & Sharpness \\\midrule

{UNet++}
& N/A & \textbf{0.975} & N/A
& N/A & N/A & N/A \\

{Bayesian UNet++}
& \textbf{0.023} & 2.237 & \textbf{0.291}
& \textbf{0.015} $(\downarrow 34.8\%)$ & 2.298 $(\uparrow 2.7 \%)$& \textbf{0.274} $(\downarrow 6.9\%)$ \\

{Dropout (40\%)}& 0.131 &0.993 &0.853 
&0.035  $(\downarrow 73.2\%)$ &\textbf{0.990}  $(\downarrow 0.3\%)$ &0.847 $(\downarrow 0.7\%)$\\

{Deep Ensemble}& 0.086 & 1.548 & 0.789 
& 0.024 $(\downarrow 70.0\%)$ & 1.366 $(\downarrow 11.8\%)$ & 0.799 $(\uparrow 1.3\%)$\\
\end{tabular*}%
}
\label{table:experiments}
\end{table}

As a result, the estimated $\mathbb{P}(Y \leq F_{X}^{-1}(p))$ by the regressor $R$ provides the calibrated probability that a random $Y$ falls into the credible interval so that we can adjust the predicted probability to the empirical probability.
\subsection{Training \& Evaluation}

\begin{figure}[b!]
\begin{subfigure}{\textwidth}
  \centering
\includegraphics[width=0.9\textwidth]{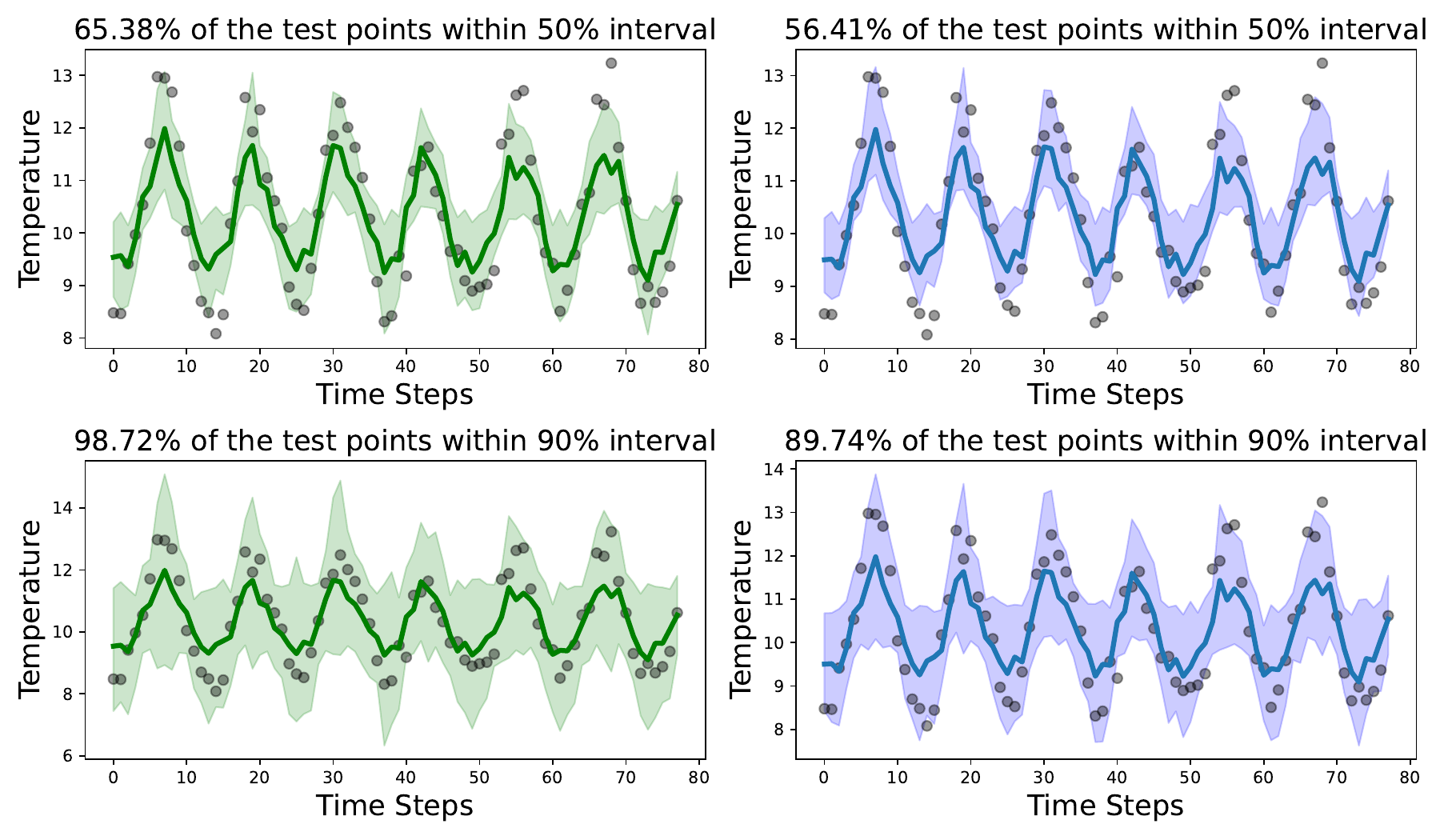}
    \label{fig:conf-90}
\end{subfigure}
\caption{$50\%$ confidence interval (Top) and $90\%$ confidence interval (Bottom) of the Bayesian UNet++ for a sample in the North West Coast of America  are given. The mean coverage percentages for confidence intervals are $63\%$ and $91\%$ for the calibrated, and $66\%$ and $98\%$ for the uncalibrated models. Thus, we choose a representative sample. Uncalibrated confidence interval plots are shown on the left (green), and calibrated plots are on the right (blue). Grey dots refer to the average temperature values for each month in the given time period (2016-2021). The percentage of the values falling within the intervals aligns more closely with the expected confidence levels, both at $50\%$ and $90\%$ in the calibrated model's plot.}
\label{fig:confidence_intervals}
\end{figure}

We train our model using the $2m$ temperature variable from 9 ensembles of the CMIP6 dataset. 1700 samples are separated for training and 100 for validation. 400 samples from the ERA5 dataset are used for fine-tuning and the construction of the calibration dataset. 116 samples from ERA5 are used in the evaluation of all methods as in \cite{unal_asan_sezen_yesilkaynak_aydin_ilicak_unal_2023}. 

\textbf{Metrics.} To measure accuracy, Mean Absolute Error (MAE) is used. MAE for calibrated models in Table \ref{table:experiments} is recalculated using the mid-quantile values from the calibrated forecaster.

Sharpness \citep{gneiting2007probabilistic} is a metric which is widely used in climate forecasting. It measures the concentration of the forecasts as
\begin{equation}
    sharpness(F_{1}, F_{2},..., F_{T}) = \frac{1}{T}\sum_{t=1}^{T}var(F_{t}).
\label{sharpness}
\end{equation}
Calibration error (CE) proposed by \cite{kuleshov2018accurate} is used for assessing the quality of the calibration of the forecasts as 
\begin{equation}
    \text{CE} (F_{1}, Y_{1}, ..., F_{T}, Y_{T}) = \frac{1}{m}\sum_{i=1}^{m}w_{j}(p_{j}-\hat{p_{j}})
\label{calib-error}
\end{equation}
where $m$ refers to the number of confidence levels $0 \leq p_{1} \leq ... \leq p_{j} \leq ... \leq p_{m} \leq 1$ and $\hat{p_{j}}$ is empirical frequency. In this setting, $w_{j}$ is chosen as 1.

\section{Results}

We use the experimental settings of \cite{unal_asan_sezen_yesilkaynak_aydin_ilicak_unal_2023}. Table \ref{table:experiments}  illustrates the impact of calibration on uncertainty quantification methods. The Bayesian model demonstrates the highest sharpness and calibration as we expected. However, there exists a trade-off between MAE and CE, with Bayesian demonstrating the lowest CE, followed by Ensemble and Dropout. Apart from the reduction in CE, we observe a decrease in MAE for Dropout and Deep Ensemble which suggests that calibration not only improves the accuracy of the network performance but also enhances the capture probability percentages of confidence intervals around point estimates. MAE is calculated using the actual $50\%$ quantile values predicted by the Isotonic Regressor.  
%

Figure~\ref{fig:confidence_intervals} demonstrates that for the calibrated case, roughly $90\%$ of the $90\%$ confidence intervals capture the true temperature values in the test dataset. We also observe the same result for $50\%$ interval. Thus, the proposed calibration produced results in line with the expected proportion of confidence intervals capturing the true outcome at the given confidence level, suggesting that the model is well-calibrated.


\begin{wrapfigure}{r!}{0.4\textwidth}
    \centering
    \includegraphics[width=0.4\textwidth]{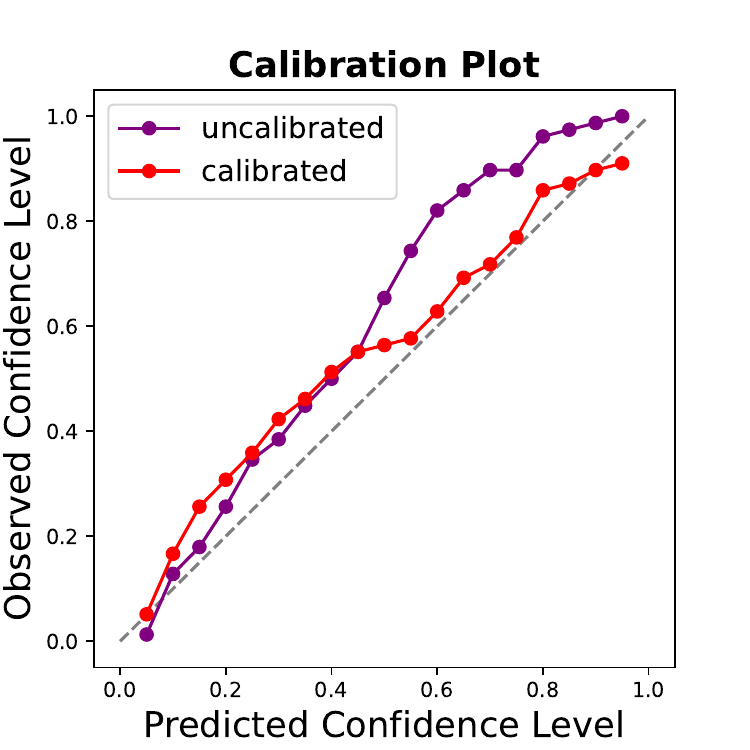}
    \caption{Calibration plot suggested by \cite{kuleshov2018accurate} given for a sample in the grid in Figure \ref{fig:confidence_intervals} to evaluate the calibration of the forecasts.
        Each predicted confidence level is plotted against its corresponding expected confidence level. Predictions illustrate the frequency of observing an outcome $Y_{t}$ at each level. We expect calibrated models to be closer to $y=x$.}
        \label{calibration_plot}
\end{wrapfigure}

CE is visualized in Figure~\ref{calibration_plot}. Equation \ref{calib-error} is applied to the values calculated for the calibration plot for each quantile, and the mean is used as the CE of that sample. After the calibration, our model converges to the $y=x$ line which indicates that the predicted confidences for the samples are closer to expected confidences, especially for quantiles larger than $50\%$.  





\section{Conclusion}
We proposed a method to enhance the sharpness and reliability of weather forecasts by calibrating them using a CDF-based calibration approach. This involved transforming the final layers of UNet++ to Bayesian. Periodically stacked multi-dimensional time-series data used as input. As we designed the output of the network to produce a CDF, we trained an isotonic regressor to calibrate the confidence intervals. We benchmarked the calibrated and uncalibrated results of three uncertainty quantification methods. Furthermore, we show that calibrating Dropout and Deep Ensemble might increase the accuracy of the network along with improving the uncertainty quantification. 

This work emphasizes the significance of calibrating neural networks while suggesting potential improvements for forecast reliability. Various fields in climate sciences can benefit from calibration since uncertainties arise from incomplete modeling of the earth and the inherent complexity of climate systems. While our focus was on temperature forecasting, this approach can be extended to predicting other essential climate variables such as precipitation, pressure, and wind components.







\bibliography{iclr2024_conference}
\bibliographystyle{iclr2024_conference}


\end{document}